\documentclass[runningheads]{llncs}

\usepackage{eccv}

\usepackage{eccvabbrv}

\usepackage{graphicx}
\usepackage{booktabs}

\usepackage[accsupp]{axessibility}  % Improves PDF readability for those with disabilities.

\usepackage{hyperref}

\usepackage{orcidlink}

\usepackage{array}
\usepackage{marvosym}
\usepackage{multirow}
\usepackage{colortbl}

\definecolor{MyBlue}{RGB}{0, 120, 215} % 蓝色
\definecolor{MyOrange}{RGB}{255, 165, 0} % 橙色

\begin{document}

\title{EAS-SNN: End-to-End Adaptive Sampling and Representation for Event-based Detection with Recurrent Spiking Neural Networks} 
\titlerunning{Event-based Sampling and Detection with SNNs}

\author{Ziming Wang\inst{1}\orcidlink{0000-0003-4270-7760} \and
Ziling Wang\inst{1}\orcidlink{0009-0007-3390-0960} \and
Huaning Li\inst{1}\orcidlink{0009-0000-4589-9409} \and
Lang Qin\inst{1}\orcidlink{0000-0002-9660-6378} \and
Runhao Jiang\inst{1}\orcidlink{0009-0001-5342-2974} \and
De Ma\inst{1,2}\orcidlink{0000-0001-8700-938X} \textsuperscript{\Letter} \and
Huajin Tang\inst{1,2}\orcidlink{0000-0003-2360-0466} \textsuperscript{\Letter}
}

\authorrunning{Z. Wang, Z. Wang, H. Li, L. Qin, R. Jiang, D. Ma, H. Tang}

\institute{College of Computer Science and Technology, Zhejiang University,
Hangzhou 310027, China \and
State Key Laboratory of Brain-Machine Intelligence, Zhejiang University,
Hangzhou 310027, China.  \textsuperscript{\Letter} Corresponding author.
\\
\email{\{zi\_ming\_wang,wzzero.hhh,lhnkevin1014,qinl,RhJiang,made,htang\}@zju.edu.cn}}

\maketitle

\begin{abstract}
  Event cameras, with their high dynamic range and temporal resolution, are ideally suited for object detection in scenarios with motion blur and challenging lighting conditions. However, 
  while most existing approaches prioritize optimizing spatiotemporal representations with advanced detection backbones and early aggregation functions, 
  the crucial issue of adaptive event sampling remains largely unaddressed. Spiking Neural Networks (SNNs), operating on an event-driven paradigm, align closely with the behavior of an ideal temporal event sampler.
  % through sparse spike communication, emerge as a natural fit for addressing this challenge.  In this study, we discover that the neural dynamics of spiking neurons align closely with the behavior of an ideal temporal event sampler. 
  Motivated by this, we propose a novel adaptive sampling module that leverages recurrent convolutional SNNs enhanced with temporal memory, facilitating a fully end-to-end learnable framework for event-based detection. Additionally, we introduce Residual Potential Dropout (RPD) and Spike-Aware Training (SAT) to regulate potential distribution and address performance degradation encountered in spike-based sampling modules. Empirical evaluation on neuromorphic detection datasets demonstrates that our approach outperforms existing state-of-the-art spike-based methods with significantly fewer parameters and time steps. For instance, our method yields a 4.4\% mAP improvement on the Gen1 dataset, while requiring 38\% fewer parameters and only three time steps. Moreover, the applicability and effectiveness of our adaptive sampling methodology extend beyond SNNs, as demonstrated through further validation on conventional non-spiking models. 
  Code is available at 
  \href{https://github.com/Windere/EAS-SNN}{Github}.

\end{abstract}

\section{Introduction}
\label{sec:intro}
% The event camera is a kind of bio-inspired sensor, that presents a new dynamic visual imaging paradigm. Different from the traditional camera that captures the illumination with a global clock synchronously, it asynchronously captures the logarithmic change of light intensity for each pixel in the dynamic scene independently. Such a change is referred to as an event. Event cameras thus feature high temporal resolution (>10K fps), high dynamic range (>120 dB), and low power consumption (<10 mW). These advantages make event cameras ideal for object detection in scenarios with high speed, challenging lighting conditions, and limit power requirements. However, the asynchronous and sparse characteristics of event cameras challenge the traditional dense network technology.
Event cameras, inspired by biological vision systems, present a transformative approach to dynamic visual sensing. Diverging from the conventional imaging model, where cameras capture scene illumination synchronously via a global shutter, event cameras operate on an asynchronous basis, logging logarithmic intensity changes at the pixel level.  Each significant change in this context is referred to as an "event". This distinct operational paradigm grants event cameras extraordinary capabilities:
% introduce a novel paradigm in dynamic visual sensing. 
% Unlike conventional cameras that synchronously capture scene illumination using a global shutter mechanism, event cameras asynchronously record logarithmic changes in light intensity at the pixel level, with each significant change denoted as an "event". This unique operational mode endows event cameras with 
exceptional temporal resolution (>10k fps), high dynamic range (>120 dB), and remarkably low power consumption (<10 mW) \cite{4444573, serrano2013128, finateu20205, gallego2020event}. These attributes render event cameras exceptionally suited for object detection under challenging environments characterized by rapid motion, extreme lighting conditions, or stringent energy constraints. 

% Nevertheless, the inherent asynchronous nature and sparsity of events pose significant challenges to conventional dense neural network architectures, which are typically designed to process data in a synchronous and uniform manner.
However, the inherent asynchronous and sparse characteristics of data captured by event cameras present significant challenges for traditional dense neural networks, which are primarily designed to process synchronous, uniformly structured data. To bridge the asynchronous event stream with the dense tensor formats utilized in ANNs, most existing frameworks incorporate a sampling-aggregation mechanism, also referred to as event representation \cite{lagorce2014asynchronous,gehrig2019end, cannici2019asynchronous,gehrig2023recurrent,liu2023motion} or event embedding. This integration effectively alleviates the computational demands engendered by the high temporal resolution of event data.  Building upon this foundation, researchers have delved into exploiting the unique spatiotemporal attributes of event data through refining event representations \cite{lagorce2016hots,sironi2018hats, li2022asynchronous,li2021graph,cordone2022object} and enhancing the design of detection architectures \cite{messikommer2020event,perot2020learning,li2022asynchronous,su2023deep,schaefer2022aegnn,gehrig2023recurrent,hamaguchi2023hierarchical}, thereby boosting network efficacy. Additionally, the aggregation module is integrated with downstream feature extraction components into a unified end-to-end training framework \cite{gehrig2019end}. Regarding event-based sampling, several studies have endeavored to enhance the sampling module from a rule-based or search-based perspective \cite{li2022asynchronous,liu2023motion,zubic2023chaos,cao2024beef}. Nonetheless, these initiatives often suffer from the lack of high-level semantic information or are hindered by artificial constraints and suboptimal search efficiency.
So far, the quest for a differentiable, adaptive sampling module that facilitates end-to-end optimization within event-based detection frameworks remains unfulfilled. 

On the other hand, SNNs, composed of biologically
plausible spiking neurons, support asynchronous event-driven computation on neuromorphic hardware\cite{akopyan2015truenorth,davies2018loihi,pei2019towards} and so present a high potential for fast inference and low power consumption. This inherent compatibility with event data makes SNNs ideally suited for event-based vision tasks.
% This positions SNNs as inherently compatible with event data. 
% Therefore, it is a kind of sparse neural network that is naturally aligned with asynchronous event-based vision data. 
Moreover, a spectrum of low-level representations \cite{sironi2018hats,lagorce2016hots,gehrig2019end} 
% conceived to facilitate the conversion from asynchronous events to dense neural network formats 
can be encapsulated within a unified SNN layer \cite{kugele2021hybrid}. Nevertheless, the challenge of training deep SNNs under long time steps persists, despite rapid advancements in spike-based learning algorithm \cite{wu2018spatio, shrestha2018slayer,zenke2021remarkable,fang2021incorporating,li2021differentiable, zhu2022training, Wang2023AdaptiveSG}. 
This makes it difficult to align the discretization window of SNNs with event response time, culminating in temporal information loss. Consequently, the current event-based detection models based on SNNs exhibit a notable performance gap when compared to their ANN counterparts. 

\begin{figure}[t]
    \centering
    \includegraphics[width=\linewidth]{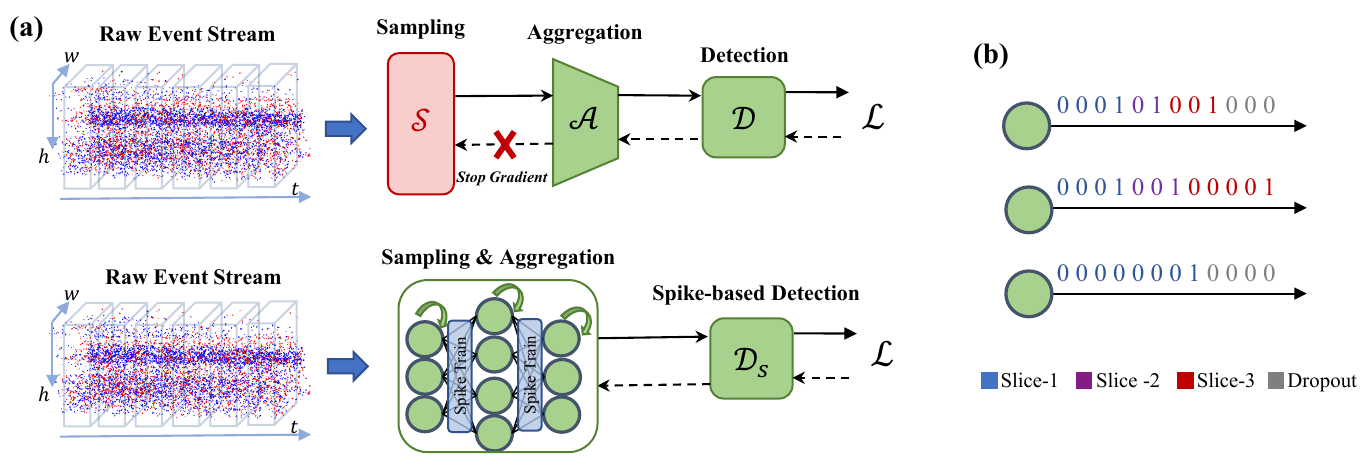}
    \vspace{-17pt}
    \caption{\textit{Left:} Implementing the end-to-end optimization for event-based sampling via recurrent SNNs. The typical rule-based methods truncate backpropagation, whereas our adaptive sampling module ensures continuous gradient flow.
    \textit{Right:} The variability and heterogeneity in sampling time and count among different spiking neurons. 
        % The upper shows the backpropagation is truncated without optimization restricted by existing rule-based methods while the proposed adaptive sampling module maintains the gradient flow. 
    }
    \label{fig.mov}
    % \vspace{-6pt}
\end{figure}

To address these challenges, this work introduces ARSNN, a learnable adaptive sampling methodology with recurrent convolutional SNNs (\cref{fig.mov}a), enabling a comprehensive end-to-end learning paradigm for event representation and feature extraction. This methodology utilizes pixel-wise spiking neurons in the sampling module to capture the dynamic information delivered by events across receptive fields. Upon the information integration surpassing a predefined threshold, the events between the last and the current spike firing times are sampled as shown in \cref{fig.mov}b. By optimizing the spike firing time and membrane potential distribution, the sampling module can be trained and coordinated with downstream spike-based detectors, thereby greatly mitigating the temporal alignment loss between the raw event stream and SNNs. 
% Moreover,
% compared with traditional event representation method, 
% the proposed SNN-based approach is optimally positioned for deployment on neuromorphic hardware, potentially curtailing overhead from data migration. 
The main contributions are outlined as follows:
\begin{enumerate}
    \item We first propose a learnable adaptive sampling method and implement an end-to-end event-based detection framework via SNNs, integrating the processes of sampling, aggregation, and feature extraction.
    \item We develop the techniques of Residual Potential Dropout (RPD) and Spike-Aware Training (SAT) to mitigate performance degradation in SNN-based sampling modules with the recurrent convolutional architecture.
    \item Empirical evaluations on neuromorphic datasets (N-Caltech 101, Gen1, and 1Mpx) demonstrate that the proposed method outperforms state-of-the-art spike-based methods with $5.85\times$ less energy consumption compared to ANNs. Furthermore, the effectiveness of our adaptive sampling method is verified across dense detection models.
\end{enumerate}

% 采样器中输出层脉冲神经元对事件相机空间感受野中的每一个像素的信息流进行捕捉，
% 当来自电流整合的信息量累积超过一定阈值时，即采样上次脉冲发放时间与此次脉冲发放时间之间的子流。
% 通过优化脉冲发放和基于膜电压的聚合，采样模块能够与下游基于SNN的检测器协同训练，从而极大地弥补了 temporal information loss between raw event stream and spiking neural networks. 此外，相比于传统的事件表征方法，基于脉冲神经网络的表征方法有望于应用在神经形态芯片上，从而降低数据迁移开销。

% 然而，因为长时间步下的深层SNN存在挑战，尽管近年来训练算法得到了快速的发展，这使得SNN的仿真时间步和事件流采样时间步难以保证一致，因而带来了时域降采样过程中的信息损失。因此，相比于ANN同行，当前基于SNN的检测方法表现出了相当的性能差距。

% In recent years, with the development of surrogate gradient learning \cite{wu2018spatio,shrestha2018slayer}

% 脉冲神经网络，被认为

% 事件相机是一种生物启发的视觉传感器，展示一种全新的动态视觉成像范式。
% 不同于传统相机同步地捕捉感受野范围内所有像素的光强， 其通过每个像素独立、异步地捕捉动态场景中的光强对数变化，这种改变被称为事件。事件相机，因而具备了高时间分辨率 (>10K fps), 高动态范围(>120 dB)，和低功耗 (<10 mW)的特点。这些优势使得事件相机非常适合应用于高速，挑战光照条件和高功耗要求的场景下的目标检测。然而事件相机异步、稀疏的特性对传统的密集网络技术提出了挑战。
% This document serves as an example submission to ECCV \ECCVyear{}.
% It illustrates the format authors must follow when submitting a paper. 
% At the same time, it gives details on various aspects of paper submission, including preservation of anonymity and how to deal with dual submissions.
% We advise authors to read this document carefully.

% The document is based on Springer LNCS instructions as well as on ECCV policies, as established over the years.
\section{Related Work}

\subsection{Spiking Neural Networks}
Spiking Neural Networks (SNNs) have raised substantial interest due to their asynchronous computation paradigm and promising energy efficiency. However, optimizing SNNs poses challenges due to the binary nature of spike firing. The advancements in backpropagation techniques for SNNs, including time-based \cite{mostafa2017supervised,zhang2020temporal,mirsadeghi2021stidi, zhang2021rectified, zhu2024exploring} and activation-based methods \cite{wu2018spatio, shrestha2018slayer,zenke2021remarkable,guo2022loss,chowdhury2022towards,li2021differentiable,Wang2023AdaptiveSG},  estimate gradients for firing time and spike generation successfully. Additionally, various spiking neuron models \cite{fang2021incorporating,guo2022reduce, yao2022glif,lian2023learnable, chen2023unleashing, hao2024a} have been proposed to enhance network heterogeneity and performance. To maintain gradient flow in deep SNNs, specialized network modules \cite{zheng2020going, duan2022temporal,fang2021incorporating} and architectures \cite{fang2021deep,su2023deep,kim2022neural, yao2021temporal,zhou2022spikformer} have been developed considering temporal dynamics in spiking neurons. These collective efforts are narrowing the performance gap between SNNs and ANNs in various tasks, including object detection \cite{wang2023toward, su2023deep, yuan2024trainable}, action recognition \cite{george2020reservoir, liu2021event, dong2024bullying10k}, point cloud recognition \cite{ren2024spiking}, and automatic speech recognition \cite{wu2020deep}.

\subsection{Event-based Detection}
In event-based detection, methodologies are bifurcated into two principal categories: dense and sparse representations. Dense representations typically convert event data into frame-based formats, subsequently processed by conventional detection algorithms \cite{lagorce2014asynchronous,cannici2019asynchronous,Cannici2020Matrix,gehrig2023recurrent,liu2023motion}. This approach, while facilitating the use of pre-trained models within an event-based context~\cite{messikommer2022bridging}, compress the inherent spatiotemporal richness of events, necessitating sophisticated event-to-frame conversion techniques.  To extract the spatiotemporal information more effectively, researchers have integrated the recurrent connections \cite{perot2020learning,li2022asynchronous,gehrig2023recurrent} and attention mechanisms \cite{wang2023dual,li2022retinomorphic}.  Recent advancements \cite{gehrig2023recurrent,liu2023motion,hamaguchi2023hierarchical}
strive to minimize detection latency without substantial performance degradation, 
through the strategic redesign of network backbones and event aggregation modules.
% thus enhancing adaptability to scenarios demanding high-speed processing.

Conversely, sparse representations, embodied by Graph Neural Networks (GNNs) and SNNs, offer substantial energy efficiency due to their inherent sparse computational models. GNN models \cite{li2021graph,schaefer2022aegnn} construct dynamic spatio-temporal graphs from event stream, updating node and edge states with incoming events. In parallel, SNN methods  \cite{cordone2022object,yuan2024trainable,su2023deep}, characterized by the internal temporal dynamics and spike-driven computation, naturally complement the asynchronous nature of event stream. 
However, given the high temporal resolution of event cameras, event-based sampling is imperative to mitigate computational demands and enhance network generalization capabilities, even for sparse representation.

% For both types of representation, due to the high temporal resolution of the event camera, the event-based sampling is needed to reduce the computational overhead and enhance the network generalization ability. 

For event-based sampling, some works have emerged to refine rule-based strategies by considering the instantaneous rate of event generation \cite{li2022asynchronous,kugele2023many,liu2023motion}. Nevertheless, these strategies frequently encounter challenges associated with hyperparameter sensitivity and the difficulty of maintaining an optimal balance between sparse and ambiguous textures in complex scenarios.
Nikola \etal \cite{zubic2023chaos} explored the optimization of sampling windows based on the Gromov-Wasserstein discrepancy from raw event data to their representations. Nonetheless, this technique has yet not fully integrated high-level semantic guidance. Cao \etal \cite{cao2024beef} proposed using SNNs for global event triggering, advocating for a feedback-update mechanism that employs pseudo labels generated by downstream detectors to guide SNN training. Despite these advancements, the integration of ANNs and SNNs remains disjointed, bridged solely by pseudo labels, which introduces artificial constraints on the optimization landscape and compromises search efficiency. In this work, we investigate an adaptive event sampling strategy leveraging learnable SNNs, marking a step towards achieving comprehensive end-to-end event-based representation learning.

\section{Methodology}
\subsection{Event-based Detection}
Consider an event stream denoted as $\mathbf{E}=\{\left(x_i, y_i, p_i, t_i\right)\}_{i\in\mathbb{N}}$, where each event $e_i$ is characterized by its spatial coordinates $x_i\in[1, W]$ and $y_i\in[1, H]$, a polarity value $p_i\in \{-1, 1\}$ indicating the change in pixel intensity, and a timestamp $t_i\in[0, +\infty)$ marking the occurrence time of events (typically measured in microseconds). 
% The dimensions $H \times W$ define the spatial resolution of the event camera, while $T$ represents the maximum observable time window, typically measured in microseconds. 
In event-based detection, annotations at a specific target time $t$ include object categories $l_j$ and associated ground truth bounding boxes $(x_j, y_j, w_j, h_j)$, encapsulated as $\mathbf{B}_{t}=\{(x_j, y_j, w_j, h_j, l_j, t)\}_{j\in\mathbb{N}}$. Ideally, a real-time detector $\mathcal{D}$ is capable of delivering predictions at the timestamp level:
\begin{equation}
\
\mathcal{D}(\mathbf{E})=\{\mathcal{D}(\mathbf{E}, t)\}_{t \ge 0}=\left\{\mathcal{D}\left(\left\{e_i\right\}_{t_i<t}\right)\right\}_{t \ge 0}
\end{equation}
Nonetheless, given the high temporal resolution of event cameras, it is challenging for event-based object detectors to process and interpret $10^4-10^5$ times per second.  
% The substantial computational demand complicates the development of efficient models capable of extracting features directly from the raw event data in a manner that is both learnable and practical. 
Moreover, the necessity for detecting moving objects at microsecond intervals is often not critical, as object motion typically occurs at a slower pace than the pixel response rate of event cameras~\cite{perot2020learning}. Consequently, most works \cite{li2022asynchronous,gehrig2023recurrent,su2023deep,yuan2024trainable} frequently incorporate a sampling operation $\mathcal{S}$ and an aggregation operation $\mathcal{A}$ to reduce the volume of sparse and asynchronous event data, thereby facilitating more efficient processing prior to the detection phase:
\begin{equation}
    \hat{\mathcal{D}}\left(\mathbf{E}, t\right) = (\mathcal{D} \circ  \mathcal{A} \circ \mathcal{S})(\mathbf{E}, t) = (\mathcal{D} \circ  \mathcal{A} \circ \mathcal{S}) (\{e_i\}_{t_i<t})
\end{equation}
By employing fixed window sampling as an example, $\mathcal{S}$ can be expressed as:
\begin{equation}
% \Sigma
   \mathcal{S}_{\Delta t, T} (\{e_i\}_{t_i<t})= \Omega_{\Delta t, T}=\{\{e_i\}_{t-(j+1)\Delta t \le t_i < t-j\Delta t}\}_{j \in [0,T/\Delta t]}
\end{equation}
where $T$ denotes the global window for sampling, and $\Delta t$ represents the interval of sampling slices.
% Assuming that $T$ can be evenly divided by $\Delta t$, the number of sampling steps, $n$, is given by $n = T/\Delta t$. 
We designate the index set that meets the criteria for fixed window sampling within the $j$-th step as $\mathbf{S}_j$. For the $j$-th step sampling,
% 即给定一段长事件流，理想的事件点是什么
% where $T_w$ is the global time window for sampling. $\Delta t$ is the sampling slice. Assuming $ t_w $ can be divided exactly by
%  $ \Delta T $, $ n = t_w/\Delta t $ is the event sampling step. We define the set of events satisfying the requirement for fixed window sampling ${t-(j+1)\Delta t \le t_i\le t-j\Delta t}$ at the $j$-th step as $\mathbf{S}_j$. 
a typical aggregation function, $\mathcal{A}^j$, based on event count histograms, is defined as follows:
  \begin{equation}
     \mathcal{A}^j (\{{e_i}\}_{i \in \mathbf{S}_j}) = {f}_j(x,y,p)= \sum_{i\in \mathbf{S}_j} \delta\left(x-x_i, y-y_i, p-p_i\right) 
     \label{eq.count}
 \end{equation}
 % \begin{equation}
 %     \mathcal{A} (\{\{{e_i}\}_{i \in \mathbf{S}_j}\}_{j\in[1,n]}) =\left\{ {f}_j(x,y,p)\right\}_{j \in [1,n]}= \left\{\sum_{i\in \mathbf{S}_j} \delta\left(x-x_i, y-y_i, p-p_i\right) \right\}_{j \in [1,n]}
 %     \label{eq.count}
 % \end{equation}
 Here, $\delta(x,y,z)$ represents the Dirac delta function, which equals 1 only when $x=y=z=0$.
 % Notably, the sampling and aggregation mechanisms, $\mathcal{S}$ and $\mathcal{A}$ respectively, can be customized for different event streams, enhancing the framework's flexibility and robustness.
 By employing this approach, the asynchronous and sparse event data are effectively down-sampled into a compact form that significantly alleviates the computational burden on the detector. 
 However, existing sampling techniques often operate independently of downstream object detectors
 % or 
 % multi-scale feature extractors
 , necessitating extensive hyperparameter tuning and showing limited adaptability. 
 % In this work, we investigate an adaptive event sampling strategy leveraging learnable SNNs, marking a step towards achieving comprehensive end-to-end event-based representation learning.
 \subsection{Reccurent Convolutional SNNs for Adaptive Event Sampling}
 \begin{figure}[t]
    \centering
    \includegraphics[width=1\linewidth]{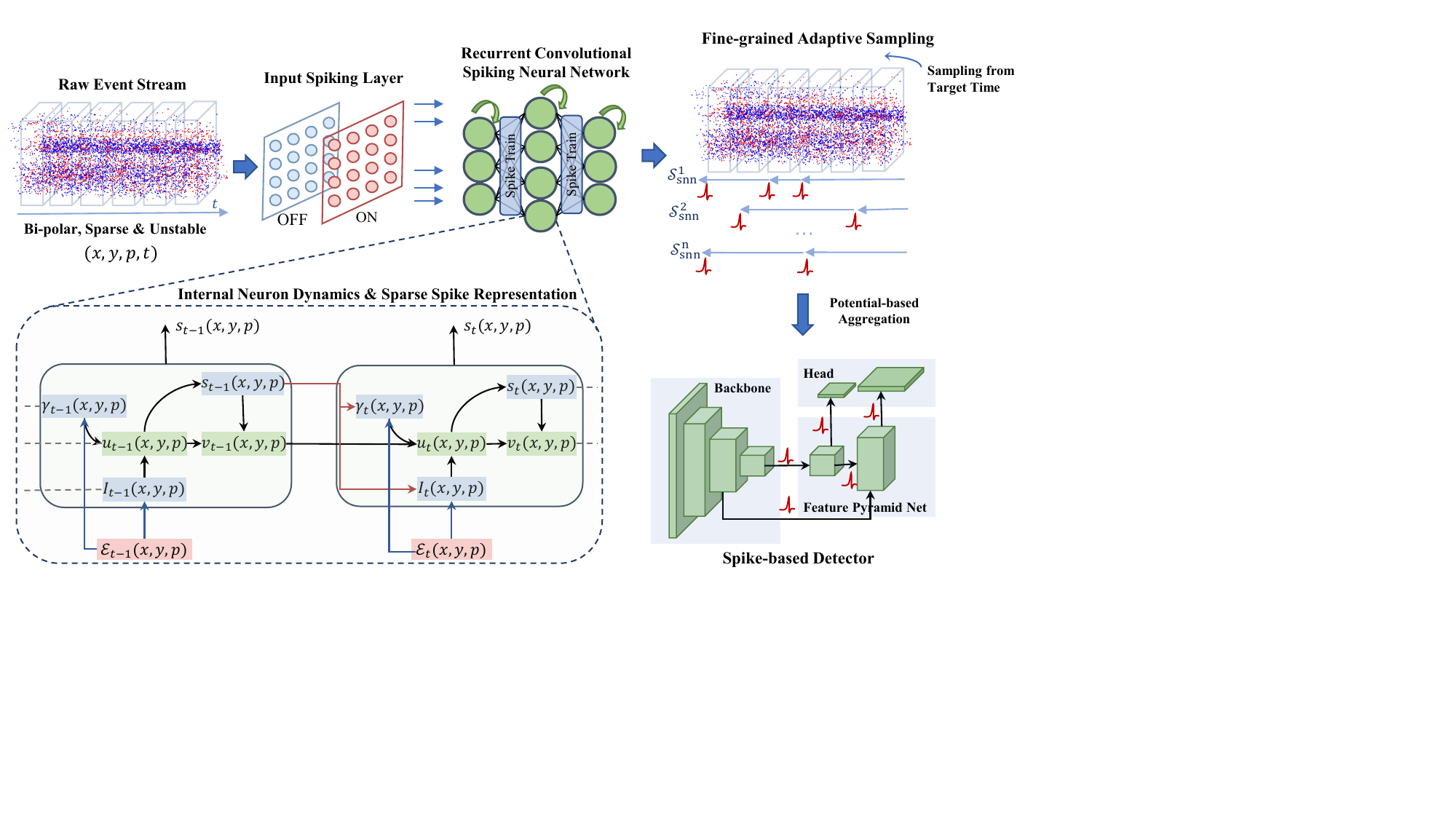}
            \vspace{-13pt}
    \caption{
    % The overall pipeline of the proposed approach. The adaptive sampling module, equipped with enhanced memory via the recurrent synapse, bridges the downstream spike-based detector and input event stream. In the unrolled graph of a spiking neuron, the   
    % solid black denotes the internal dynamics of the spiking neuron. The blue and red solid line represent feedforward and recurrent connections, respectively.
    Illustration of the overall pipeline. The adaptive sampling module, augmented with recurrent synaptic connections for improved memory mechanism, bridges the input event stream with the downstream spike-based detector. Within the unrolled computation graph, internal neuron dynamics are indicated by solid black lines, while blue and red solid lines depict the feedforward and recurrent pathways, respectively.
    }
        % \vspace{-9pt}
    \label{fig.pipeline}
\end{figure}
 In this study, our goal is to develop a learnable sampling module (\cref{fig.pipeline}), denoted as $\mathcal{S}_\text{SNN}$, utilizing SNNs to address the challenge of non-differentiable, end-to-end learning that constrains rule-based sampling methods. 
 % In this work, we aim to implement a learnable sampler $\mathcal{S}_\text{SNN}$ using SNNs to solve the problem of non-differential, end-to-end learning of traditional rule-based samplers.
% The leaky-integrate-fire (LIF) spiking model is widely adopted in neuromorphic computing for its low computational overhead which can be formalized as follows:
% 膜电压在接收到输入脉冲时会累积输入电流，反之会按照固定的decay进行衰减
% \begin{equation}
%     \tau_m \frac{d {u}^\ell_i(t)}{d t}=-{u}^\ell_i(t)+R {I}^\ell_i(t) \\
% \end{equation}
% The discrete form of the LIF model via the first-order Euler method, with a time slice of $\Delta t_s$, is given by:

\noindent {The Leaky-Integrate-and-Fire (LIF) model}, popular in neuromorphic computing due to its minimal computational requirements.
% , forms the foundation for our approach. 
The discrete LIF equations, via the first-order Euler method with a time slice of $\Delta t_s$, are given by:
% \vspace{-1pt}
\begin{equation}
% \vspace{-1pt}
\begin{gathered}
u_i^{\ell}[t]=g\left(v_i^{\ell}[t-1], I_i^{\ell}[t]\right)=\left(1-\frac{\Delta t_s}{\tau_m}\right) v_i^{\ell}[t-1]+\frac{\Delta t_s}{\tau_m} I_i^{\ell}[t] \\
s_i^{\ell}[t]=\Theta\left(u_i^{\ell}[t]-\theta\right)= \begin{cases}1, & u_i^{\ell}[t] \geq \theta, \\
0, & u_i^{\ell}[t]<\theta\end{cases} \\ \label{eq.spike}
v_i^{\ell}[t]=\left\{\begin{array}{l}
u_i^{\ell}[t]-\theta \cdot s_i^{\ell}[t], \text { soft reset } \\
u_i^{\ell}[t] \cdot\left(1-s_i^{\ell}[t]\right)+u_{\text {reset }} \cdot s_i^{\ell}[t], \text { hard reset }
\end{array}\right.
\end{gathered}
\end{equation}
%     \begin{gather}
% {u}^\ell_i[t]= g({v}^\ell_i[t-1],I^\ell_i[t]) = (1-\frac{\Delta t_s}{\tau_m}){v}^\ell_i[t-1]+\frac{\Delta t_s}{\tau_m} I^\ell_i[t] \\
% s^l_i[t] = \Theta(u^\ell_i[t]-\theta)=\begin{cases}
% 1, \ \ \ u^\ell_i[t] \ge \theta, \\
% 0, \ \ \  u^\ell_i[t] < \theta 
% \end{cases} \label{eq.spike}\\
% {v}^\ell_i[t]=\begin{aligned}\left\{\begin{array}{l} 
% u^\ell_i[t]-\theta \cdot s^\ell_i[t], \text { soft reset}\\
% u^\ell_i[t] \cdot(1-s^\ell_i[t])+u_{\text {reset }} \cdot s^\ell_i[t], \text { hard reset} 
% \end{array}\right.
% \end{aligned}
% \end{gather}
Here, ${u}^\ell_i[t]$ and $I^\ell_i[t]$ denote the membrane potential and input current of the $i$-th neuron in the $l$-th layer, respectively, with $\tau_m$ as the membrane time constant. 
Upon receiving spike inputs, a neuron accumulates the input current $I^\ell_i[t]$ into its membrane potential ${u}^\ell_i[t]$, which decays at a constant rate. When the membrane potential ${u}^\ell_i[t]$ exceeds the threshold $\theta$, a spike is emitted, resetting ${u}^\ell_i[t]$ to $v^\ell_i[t]$. The LIF dynamic closely aligns with the desired properties of event-driven sampling, where events should be selectively sampled as the accumulated information from the asynchronous event stream reaches a threshold. Consequently, we propose to associate spike firing in SNNs with event-based sampling, integrating this functionality into an end-to-end optimization framework. 

% A subset of events should be selectively sampled as the accumulated information from the asynchronous event stream reaches a certain threshold. Consequently, we associate the spike firing in SNNs with event-based sampling, integrating this functionality into an end-to-end optimization framework. 
Firstly, restricted by the poor support for sparse spike-driven computation on GPUs, we perform early aggregation using event count as shown in \cref{eq.count} with a small time slice $\Delta t_m$ and obtain the actual $T_m$-step inputs for  $\mathcal{S}_\text{SNN}$. 
Following early aggregation, a distinct spiking neuron is allocated to each spatial position $(x, y)$ and polarity $p$, with each neuron exhibiting unique behavior by accumulating its spatially localized input over time:
% After early aggregation, each spiking neuron $\mathcal{SN}$ is used for each position $(x,y)$ and each polarity $p$, and the corresponding spatially local input response is accumulated over time:

% It is worth noting that in general, the closer the timestamps of events are to the detection target timestamp $t^\prime$, the greater the impact on the detection result at $t^\prime$. Therefore, in this case, the input closer to $t^\prime$ in time is preferentially sent to $\mathcal{S}_\text{SNN}$. 
% % In the actual streaming process, the stack can be used to implement such first-in-last-out computing logic.
% This operation is also consistent with the temporal information dynamics in the SNNs which has been observed that information becomes highly concentrated in earlier few timesteps \cite{kim2023exploring}.
% \vspace{-5pt}
\begin{equation}
\begin{aligned}
    I_t(x,y,p) =  (W_\text{conv}&  * f_t)(x, y,p) + b_\text{conv}(p) \\
u_t(x,y,p)= g(I_t(x,y,p), v_t&(x,y,p)) = \gamma\cdot v_t(x,y,p)+I_t(x,y,p)
\end{aligned}
\label{eq.snn_samp}
\end{equation}
where $\gamma = 1-\frac{\Delta t_s}{\tau_m}$ denotes a constant decay factor. 
Our empirical findings suggest that an effective sampler must possess an advanced temporal credit assignment capability to extract informative and discriminative substreams from the inherently noisy and fluctuating spatiotemporal event data.
% In practice, we find that an effective sampler needs the enhanced ability of temporal credit assignment to extract informative and discriminative substream from the noisy and unstable spatiotemporal event stream.
Therefore, we incorporate recurrent synaptic connections, as depicted in the bottom left portion of \cref{fig.pipeline}, to enhance the temporal representational capacity of SNNs:
% \vspace{-\upl}
\begin{equation}
% \vspace{-\upl}
\begin{aligned}
u_t(x,y,p) &= \gamma_t(x,y,p) \cdot v_{t-1}(x,y,p)+I_t(x,y,p) \\
 I_t(x,y,p) &=  (W^{f}_{I}  * f_t)(x, y,p) +   (W_{I}^{s}  * s_{t-1})(x, y,p) +b_I(p)\\
  \gamma_t(x,y,p) &= \sigma \left((W^{f}_{\gamma}  * f_t)(x, y,p) +  (W_\gamma^{s}  * s_{t-1})(x, y,p) +b_\gamma(p) \right) 
\end{aligned}
\end{equation}
Here, the spike firing $s_{t-1}(x,y,p)$ at the last time step contributes to the input current $I_t(x,y,p)$ and decay factor $\gamma_t(x,y,p)$ at the current time step directly through recurrent synapse $W_{I}^{s}$ and $W_\gamma^{s}$. $\sigma$ denotes the sigmoid function which normalizes the decay factor $\gamma_t$.
We refer to SNNs after this modification as “RSNN”.
When the spike is emitted with over-threshold membrane potential, the event slice between the last spike firing time $t^{k-1}(x,y,p)$ and the current spike firing time $t^{k}(x,y,p)$ is sampled, so as to achieve local sampling:
% \begin{equation}
% \begin{gathered}
% \Omega^k(x, y, p)=\left\{\left(x_i, y_i, p_i, t_i\right) \mid t_i \in\left(t^{k-1}(x, y, p), t^k(x, y, p)\right] \&[x, y, p]=\left[x_i, y_i, p_i\right]\right\} \\
% \text { where } \quad s_{t^k}(x, y, p)=1 \& t^{k-1}(x, y, p)<t^k(x, y, p) \tag{8}
% \end{gathered}
% \end{equation}
% \begin{equation}
% \begin{split}
% \Omega^k(x, y, p) &= \left\{\left(x_i, y_i, p_i, t_i\right) \mid t_i \in \left(t^{k-1}(x, y, p), t^k(x, y, p)\right] \& [x, y, p] = \left[x_i, y_i, p_i\right]\right\} \\
% \text{where} \quad s_{t^k}(x, y, p) &= 1 \& t^{k-1}(x, y, p) < t^k(x, y, p) \tag{8}
% \end{split}
% \end{equation}
% \begin{equation}
% \begin{aligned}
%     \Omega^k(x,y,p)=\left\{(x_i,y_i,p_i,t_i)|t_i &\in ({t^{k-1}(x,y,p), t^k(x,y,p)]}  \ \& \ [x,y,p] =[x_i,y_i,p_i] \right \}  \\ 
%     \text{where}  \ \ \ s_{t^k}(x,y,p) &= 1 \ \& \ t^{k-1}(x,y,p)<t^k(x,y,p) \tag{8}
%     \end{aligned} 
% \end{equation}
% \begin{align}
% \Omega^k(x, y, p) &= \left\{\left(x_i, y_i, p_i, t_i\right) \mid t_i \in \left(t^{k-1}(x, y, p), t^k(x, y, p)\right] \& [x, y, p] = \left[x_i, y_i, p_i\right]\right\} \notag \\
% & \quad \text{where} \quad s_{t^k}(x, y, p) = 1 \& t^{k-1}(x, y, p) < t^k(x, y, p) \tag{8}
% \end{align}
% \vspace{-2pt}
\begin{multline}
% \vspace{-\upl}
\hspace*{-.84\parindent} \Omega^k(x, y, p) = \left\{\left(x_i, y_i, p_i, t_i\right) \mid t_i \in (t^{k-1}(x, y, p), t^k(x, y, p)] \& [x, y, p] = [x_i, y_i, p_i]\right\} \\ 
\text{where} \quad s_{t^k}(x, y, p) = 1 \ \&  \ t^{k-1}(x, y, p) < t^k(x, y, p) \label{eq:Omega}
\end{multline}

% \begin{multline}
% \Omega^k(x, y, p) = \left\{ (x_i, y_i, p_i, t_i) \mid t_i \in (t^{k-1}(x, y, p), t^k(x, y, p)] \& [x, y, p] = [x_i, y_i, p_i] \right\} \\
% \text{where} \quad s_{t^k}(x, y, p) = 1 \& t^{k-1}(x, y, p) < t^k(x, y, p) \tag{8}
% \end{multline}

% \begin{align}
% \Omega^k(x, y, p) &= \{ (x_i, y_i, p_i, t_i) \mid t_i \in (t^{k-1}(x, y, p), t^k(x, y, p)] \notag \\quad \& [x, y, p] = [x_i, y_i, p_i] \} \notag 
% &\quad \text{where} \ s^k(x, y, p) = 1 \& t^{k-1}(x, y, p) < t^k(x, y, p) \} \tag{8}
% \end{align}
% Importantly, as the firing times of the different spiking neurons at different spatial coordinates actually show a certain variability, event sampling also shows a high degree of flexibility.
This approach optimizes event sampling by converting it into learning proper firing times across spiking neurons at distinct spatial coordinates. 
After event-based sampling, the remained challenge is to effectively integrate each sampled set into a coherent embedding. Empirically, we find that decoupling the sampling and the aggregation modules can hinder training due to unstable gradient flow. Thus, we unify the sampling and the aggregation within a single SNN, using the potential accumulation within the specified sampling window $(t^{k-1}(x,y,p),t^k(x,y,p)]$ as the aggregation input for the downstream network:
% the neuronal membrane potential accumulation in the corresponding time period as the aggregation results for the downstream detector:
% \vspace{-\upl}
\begin{equation}
% \vspace{-\upl}
    \hat{f}_k(x,y,p)=  \sum_{(x_i,y_i,p_i,t_i) \in  \Omega^k(x,y,p)} u_{t_i}(x_i,y_i,p_i) 
    % \left\{   \sum_{t={t^{k-1}+1}}^{t^k} u_{t_i}(x_i,y_i,p_i)\ \ \  \text{where} \ \ s_{t^k}(x,y,p)=s_{t^{k-1}}(x,y,p) = 1 \ \& \ t^{k-1}<t^k  \right\}
    \label{eq.pa}
\end{equation}
Notably, we adopt the hard reset to clear the membrane potential from the previous sample.
Such a model incorporating the spike-driven sampling mechanism and the synaptic recurrent connection is referred to as the ‘ARSNN’ for the latter discussion.

% The above leaky-integrate-and-fire neuron dynamics are just consistent with the properties we expect for an event sampler, that is, an event subset should be sampled when the amount of information accumulated from the asynchronous event stream exceeds a certain threshold. Therefore, we consider associating the spike firing in SNNs with the event stream sampling and taking it into end-to-end optimization (Fig. \cref{fig.pipeline} ).

% From the neuron dynamics shown in the above equations, we 
% Accumulate the input response $I^l i[t]$for each pulse over time, and when the response reaches a certain threshold $u \text{th}$, then issue an output pulse such that the neuron behavior is consistent with the expected event sampler $\mathcal{S}$, The amount of information is continuously accumulated with the input events. When the overall amount of information reaches a certain threshold, the corresponding event stream subset is sampled. Therefore, it is natural for us to consider the combination of pulse sending behavior and event stream sampling and bind them into the optimization of the end-to-end event detector

  \subsection{Mitigating Degradation of Spike-based Sampling}
\label{sec.mitigating}
\begin{figure}[t]
    \centering
    \includegraphics[width=1\linewidth]{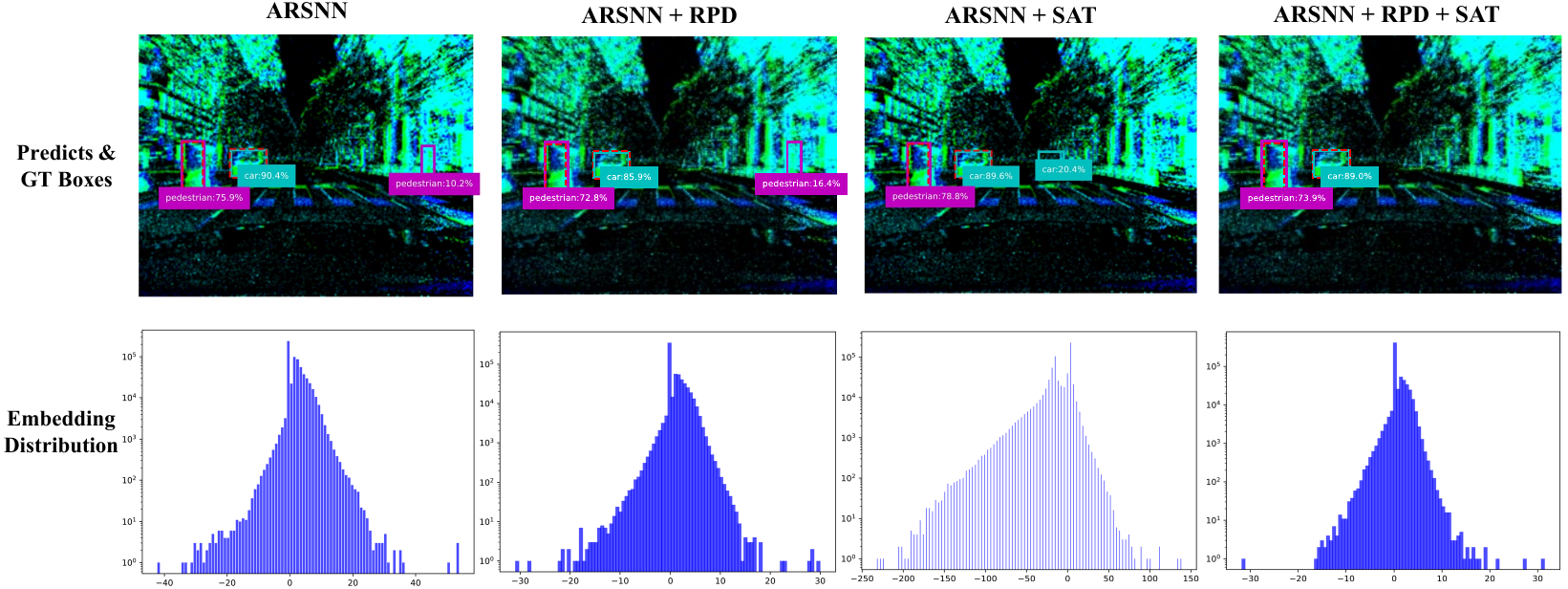}
    \vspace{-15pt}
    \caption{Visualizing the detection results and corresponding embedding distributions of different models in ablation study. The red dashed line highlights the ground truth.}
    \label{fig.embedding}
    % \vspace{-11pt}
\end{figure}
In light of the adaptive sampling, it can be integrated with the downstream detector for end-to-end training via backpropagation through time (BPTT) \cite{werbos1988generalization} with surrogate gradient functions \cite{wu2018spatio,shrestha2018slayer,fang2021incorporating}. However, this aggregation method in \cref{eq.pa} implicitly optimizes the sampling module by relying solely on the aggregation of membrane potentials, which lacks information from the firing threshold for optimizing the spike firing timing. 
Formally, we leverage the gradients from the activation-based technique in SNN training to estimate the update of firing time $t^k$
% we calculate the update of firing time $t^k$ based on the gradients from the activation-based method in SNN training 
(detailed derivation provided in the Supplementary Material):
 \begin{equation}
     % \vspace{-1pt}
     \Delta t^k(x,y,p)  \propto   -\frac{\partial \mathcal{L}((\mathcal{D} \circ  \mathcal{A} \circ \mathcal{S}
)(\mathbf{E}, t), \textbf{B}_t)}{\partial \hat{f}_k(x,y,p)}({u_{t^k}(x,y,p)-u_{t^k-1}(x,y,p))}
 \end{equation}
 This reveals that the adjustment in firing time $t^k$
  depends solely on the gradient of the membrane potential $u_{t^k}(x,y,p)$ and the instantaneous change in potential at the firing moment. It disregards the relationship between the firing time update $\Delta t^k$  and the firing threshold $\theta$.  To address this issue, we introduce Spike-aware Training (SAT), which incorporates the firing spike 
 $s_{t^k}(x,y,p)$  into the aggregation function as follows:
  \begin{equation}
       % \vspace{-1pt}
     \hat{f}_k(x,y,p)=  s_{t^k}(x,y,p) \sum_{t^\prime={t^{k-1}(x,y,p)+1}}^{t^k(x,y,p)} u_{t^\prime}(x,y,p)\ \ \   
 \end{equation}
 Given $s_{t^k}=1$, this modification does not alter the forward computation of the aggregation function. However, it effectively rewires the gradient flow:
  \begin{equation}
       % \vspace{-1pt}
     \Delta t^k  \propto  -\frac{\partial \mathcal{L}}{\partial \hat{f}_k}(1+\sum_{t^\prime} u_{t^\prime}h_\alpha(u_{t^k}-\theta))({u_{t^k}-u_{t^k-1})},\quad  \forall \ (x,y,p)
 \end{equation}
 where $h_\alpha(x)$ is the surrogate gradient function used to circumvent the non-differential problem of the Heaviside function in \cref{eq.spike}. 
 % We use the rectangular surrogate function \cite{wu2018spatio}.
  Here, $1+u_{t^k}h_\alpha(u_{t^k}-\theta)$ represents the adjustment from threshold-based gradient redirection, enhancing the update of firing time 
$s_t^k(x,y,p)$ especially when near the threshold 
$\theta$, thereby achieving more precise sampling away from critical points.
Moreover, the implementation of a spike-based sampling module introduces the dilemma of handling data after the final spike firing.
% For instance, consider a scenario where a spiking neuron emits three spikes due to sufficient information flow, with each spike corresponding to a set of events sampled (first row in \cref{fig.mov} b). The question arises: should the events arriving in the final three time steps be amalgamated into a new sampling set?
An intuitive idea is to incorporate the remaining information into the embedding to reduce information loss. However, this leads to a scenario where neurons, even without firing, accumulate substantial information for representation. This inadvertently causes sampling degradation by diminishing the incentive for the sampling module to learn accurate spike timing. Hence, we propose the strategy of Residual Potential Dropout (RPD) to exclude all events following the last spike, enhancing the learning of precise firing moments.  
To illustrate the efficacy of RPD, we show the potential aggregation distribution with and without RPD. The results in \cref{fig.embedding} reveal a more compact feature distribution with RPD.
Moreover, employing SAT for gradient guidance (as in the case of ARSNN+SAT), 
we observe that sampler training without RPD tends to exhibit mode collapse 
towards the non-firing spectrum with
concentrated negative membrane potentials. 

\section{Experiments}
The effectiveness of the proposed approach is first validated in \cref{sc.ablation}. In \cref{sc.pc}, we compare our framework with state-of-the-art event-based methods across benchmark datasets. \cref{sc.ea} explores adaptability, hyper-parameter sensitivity, and sampling diversity, while \cref{sc.ee} examines energy efficiency.

% The effectiveness of the proposed approach is initially substantiated in \cref{sc.ablation}. Subsequently, in \cref{sc.pc}, we compare our comprehensive framework with state-of-the-art event-based methods across benchmark datasets. Additionally, \cref{sc.ea} explores the adaptability, hyper-parameter sensitivity, and sampling diversity, while \cref{sc.ee} delves into its energy efficiency.

% In \cref{sc.pc}, we compare our whole framework with the state-of-the-art
% event-based approaches on benchmark detection datasets. Furthermore, we verify the scalability, hyper-parameter robustness and sampling heteroGeneity in \cref{sc.ea}, as well as energy efficiency in \cref{sc.ee}. 

\noindent {\footnotesize \textbf{{Implementation Details.}} }
% Our experiments are executed on four RTX 3090 GPUs. 
The ADAM optimizer is applied with learning rates of 1e-3 for Gen1, 1e-4 for N-Caltech 101, and 5e-4 for 1Mpx. Consistent with \cite{gehrig2023recurrent,hamaguchi2023hierarchical}, data augmentation techniques, including random Zoom-in and horizontal flipping, are employed for all datasets. For Gen-1, event sampling is capped at 200ms prior to the timestamp of labeling. For 1Mpx, we preprocess the event stream following \cite{gehrig2023recurrent}, while using a subsequence of length 4 for BPTT training.
Unlike the hard reset used in the sampling module,  downstream networks implement a soft reset strategy. Instead of employing a learning rate scheduler, we leverage model weight smoothing during training through the use of an exponential moving average with a momentum of $0.9999$. Additionally, we utilize SpikingJelly \cite{fang2023spikingjelly} to enhance the training efficiency of the spike-based backbone.
% All experiments are conducted with 4 RTX 3090 GPUs. For the spike-based backbone, we adopt SpikingJelly to help improve the training speed. The ADAM optimizer is adopted across all experiments. The learning rate is set as 1e-3 and 1e-4 for Gen1 and N-Caltech 101, respectively.  For both datasets, following \cite{gehrig2023recurrent,hamaguchi2023hierarchical}, we use the random Zoom-in augmentation and horizontal flip as data augmentation. For GEN1, to control data processing time, we restrict the events up to 200ms before the labeling time can be sampled. In contrast to the hard reset mechanism used in the sampling module, the soft reset is used in downstream networks. Instead of using the learning rate scheduler, the model exponential moving average with $m=0.9999$ is adopted to smooth model weights during training.

\noindent {\footnotesize \textbf{{Benchmark Datasets.}} }  The Neuromorphic-Caltech 101 (N-Caltech 101) dataset~\cite{orchard2015converting}, captured by an ATIS event camera \cite{posch2010qvga}, comprises 8,709 event streams, each containing 300 ms of saccades with bounding box annotations and object contours. We partition this dataset into an 8:2 split for training and validation. The Prophesee Gen1 dataset \cite{de2020large} offers over 39 hours of automotive recordings, annotated with more than 228k car and 28k pedestrian labels, facilitating event-based detection in diverse driving scenarios.
The 1Mpx dataset \cite{perot2020learning} features high-resolution driving recordings (1280×720) across various scenarios, including over 24 million labeled bounding boxes of cars, pedestrians, and two-wheelers, annotated at 60Hz. Following the original evaluation protocol, we exclude prediction boxes with a side length below 10 (20) pixels or a diagonal shorter than 30 (60) pixels after non-maximum suppression for the Gen1 (1Mpx) dataset.
% prediction boxes with either a side length below 10 (20) pixels or a diagonal shorter than 30 (60) pixels are excluded after non-maximum suppression for the Gen1 (1Mpx) dataset.
% 14 hours of driving recording, acquired 

% The Neuromorphic-Caltech 101 (N-Caltech 101) dataset~\cite{orchard2015converting} is recorded by moving ATIS event camera \cite{posch2010qvga} on the 101 object categories in three saccades. It includes 8,246 event streams, each of which contains 300 ms of saccades, as well as bounding boxes. We split the dataset into 8:2 for training and validation, respectively. The Prophesee Gen1 dataset \cite{de2020large} is the other challenging neuromorphic dataset for event-based detection in diverse driving scenarios, that includes more than 39 hours of automotive recordings with more than 228k car and 28k pedestrian annotations. 
% Following the evaluation policy in the original paper, we remove the prediction boxes with a side length of less than 20 pixels or a diagonal of less than 30 pixels after non-maximum suppression. 

% It contains open roads and very diverse driving scenarios, ranging from urban, highway, suburbs
% and countryside scenes, as well as different weather and illumination conditions.

% by mounting the ATIS sensor on a motorized pan-tilt unit and having the sensor move while it views Caltech101 examples on an LCD monitor as shown in the video below
\subsection{Ablation Study}
\label{sc.ablation}
% In this section, we compare the proposed method with the commonly used event count method \cite{fang2021incorporating,gehrig2023recurrent,su2023deep} on both Gen1 and N-Caltech 101 datasets. In addition, two embedding methods of SNN and RSNN based on the potential aggregation as shown in \cref{eq.pa} are proposed to highlight the effectiveness of each component. The results are shown in Table \ref{tb.ablation}. By comparing SNN and RSNN, we can find that recurrent synapse connections help improve the network performance effectively on both datasets (+6.2\% on N-Caltech 101 and +4.9\% on Gen1). Additionally, the results on Gen1 show that using SAT and RPD alone will bring some performance degradation. For N-Caltech 101, such problems can sometimes be more serious. For example,  the lack of RPD module caused 3.5\% mAP$_{50}$ loss, which is even worse than the result of RSNN. In contrast, 
% combining both SAT and RPD, the performance of SNNs is significantly improved by 2.7\% and 1.3\% compared to RSNN without adaptive sampling on N-Caltech 101 and Gen1 datasets, respectively 

In this section, we conduct a comparative analysis to ascertain the efficacy of individual components within the proposed sampling module, as detailed in Table \ref{tb.ablation}. In addition to comparing with Event Count used in early aggregation, we introduce two novel embedding strategies leveraging SNNs and RSNNs, predicated on potential aggregation. The results show that the integration of recurrent synaptic connections significantly enhances network performance on both datasets, with observed improvements of 6.7\% on N-Caltech 101 and 4.9\% on Gen1. Notably, employing the SAT and RPD modules in isolation with adaptive sampling sometimes results in a performance decline compared to RSNN (\textcolor{MyBlue}{58.6\%} vs. \textcolor{MyOrange}{63.4\%} in mAP$_{50}$ on N-Caltech 101). 
Conversely, when SAT and RPD are synergistically combined within the SNN framework, there is a significant performance improvement, yielding increases of 3.0\% and 1.2\% in mAP$_{50}$ over RSNN on the N-Caltech 101 and Gen1 datasets, respectively. This demonstrates the critical importance of integrating these components to achieve optimal network performance. Furthermore, Table \ref{tb.rep_cp} compares our approach with 
established event representations such as Voxel Grid \cite{zhu2019unsupervised}, Time Surface \cite{sironi2018hats}, and Voxel Cube \cite{cordone2022object}. 
It shows that the proposed adaptive method outperforms these alternatives on both datasets. Moreover, the framework can incorporate various event representations at the early aggregation stage for better performance.
\begin{table}[t]
    \centering
    \begin{minipage}{0.54\linewidth}
        \centering
                \caption{Performance comparison between Event Count and the proposed spike-based approaches, with and without the adaptive sampling mechanism. }
        \vspace{-6pt}
\label{tb.ablation}
        \resizebox{\linewidth}{!}{
      \begin{tabular}{c|ccccccc}
% \toprule
\hline
\centering
\multirow{2}{*}{{Method}} & \multirow{2}{*}{{RPD}} & \multirow{2}{*}{{SAT}} & \multicolumn{2}{c}{{N-Caltech 101}}     &  & \multicolumn{2}{c}{{Gen1}}         \\
\cline{4-5} \cline{7-8}
                                 &                               &                               & {mAP$_{50}$} & {mAP$_{50:95}$} &  & {mAP$_{50}$} & {mAP$_{50:95}$} \\ \cline{1-5} \cline{7-8} 
Event Count                      & -                             & -                             & 0.587            & 0.369                  &  & 0.611            & 0.321                  \\
SNN                              & -                             & -                             & 0.567            & 0.367                  &  & 0.670            & 0.366                  \\
RSNN                             & -                             & -                             & \textcolor{MyOrange}{0.634}            & 0.412                  &  & 0.719            & 0.399                  \\
ARSNN                            & $\times$                      & $\times$                      & 0.600            & 0.398                  &  & 0.687            & 0.373                  \\
ARSNN                            & $\checkmark$                  & $\times$                      & \textcolor{MyBlue}{0.586}            & 0.391                  &  & 0.714            & 0.393                  \\
ARSNN                            & $\times$                      & $\checkmark$                  & 0.592            & 0.376                  &  & 0.714            & 0.398                  \\
ARSNN                            & $\checkmark$                  & $\checkmark$                  & 0.664            & 0.437                  &  & 0.731            & 0.409                  \\ 
% \bottomrule
\hline
\end{tabular}
}
    \end{minipage}
    % \hspace{0.2\linewidth}
    \begin{minipage}{.45\linewidth}
        \centering
        \caption{Performance comparison with popular event representation methods on both Gen1 and N-Caltech 101 datasets.}
\label{tb.rep_cp}
        \vspace{-6pt}
                \resizebox{\linewidth}{!}{
\begin{tabular}{c|ccccc}
\hline
\multirow{2}{*}{{Method}} &  \multicolumn{2}{c}{{N-Caltech 101}} &                      & \multicolumn{2}{c}{{Gen1}}         \\
\cline{2-3}\cline{5-6} 

                                                              & {mAP$_{50}$} & {mAP$_{50:95}$} &  & {mAP$_{50}$} & {mAP$_{50:95}$} \\ 
                                                              \cline{1-3} \cline{5-6}
Event Count                                                  & 0.587             & 0.369                  &                      & 0.611            & 0.321                  \\ Voxel Grid                            & 0.598             & 0.391                  & \multicolumn{1}{l}{} & 0.515            & 0.247                  \\
Time Surface                                                & 0.497             & 0.288                  & \multicolumn{1}{l}{} & 0.654            & 0.354                  \\
Voxel Cube                                                & 0.589             & 0.401                  & \multicolumn{1}{l}{} & 0.651           & 0.340                  \\
\rowcolor{blue!9}  % 高亮第三行，使用青色
SNN                                                       & 0.567             & 0.367                  &                      & 0.670            & 0.366                  \\
\rowcolor{blue!9}  % 高亮第三行，使用青色
RSNN                                                      & 0.634             & 0.412                  &                      & 0.719            & 0.399                  \\
% ARSNN                            & $\times$                      & $\times$                      & 0.600             & 0.398                  &                      & 0.687            & 0.373                  \\
% ARSNN                            & $\checkmark$                  & $\times$                      & 0.586             & 0.391                  &                      & 0.714            & 0.393                  \\
% ARSNN                            & $\times$                      & $\checkmark$                  & 0.550             & 0.357                  &                      & 0.714            & 0.398                  \\
\rowcolor{blue!9}  % 高亮第三行，使用青色
ARSNN                                       & 0.664             & 0.437                  &                      & 0.731            & 0.409                  \\ \hline
\end{tabular}
}
    \end{minipage}
            % \vspace{-4pt}
\end{table}
\begin{table}[t]
\caption{Performance comparison with the state-of-the-art methods on the Gen1 dataset. The letters M and S in parentheses denote medium and small-scale network structures, respectively, achieved by scaling the network width.
 }
\label{tb.gen1}
\vspace{-8pt}
\resizebox{\linewidth}{!}{
\begin{tabular}{@{}lcccccccc@{}}
\hline
\multirow{2}{*}{\textbf{Method}} & \multirow{2}{*}{\textbf{Representation}} & \multirow{2}{*}{\textbf{Backbone}} & \multirow{2}{*}{\textbf{Head}}               & \multirow{2}{*}{\textbf{Temporal}} & \multirow{2}{*}{\textbf{\#Params}} & \multirow{2}{*}{\textbf{Timestep}} &   \multirow{2}{*}{\textbf{$\text{mAP}_{50}$}} &\multirow{2}{*}{\textbf{$\text{mAP}_{50:95}$}}       \\ 
                                 &                                     &                                    &                                              &                                    &                                       &                                      &  &  \\ \hline
RED \cite{perot2020learning}                              & Event Volum                         & CNN+RNN                            & SSD                                          & Yes                                & 24.1M                                 & -                                    & -                & 0.40                    \\
RVT \cite{gehrig2023recurrent}                             & HIST.                               & Transformer+RNN                  & YOLOX                                        & Yes                                & 18.5M                                & 21                                   & -                & 0.472                  \\
ERGO-12 \cite{zubic2023chaos} & ERGO-12 & Transformer & YOLOv6& No & - & - & -& 0.504 \\

AEGNN  \cite{schaefer2022aegnn}                          & Graph                               & GNN                                & YOLO                                         & No                                 & 20.0M                                  & -                                    & -                & 0.163                  \\
NVS-S   \cite{li2021graph}                         & Graph                               & GNN                                & YOLO                                         & No                                 & 0.9M                                   & -                                    & -                & 0.086                  \\
TAF    \cite{liu2023motion}                          & CNN                                 & CNN                                & YOLOX                                        & No                                 & 14.8M                                  & -                                    & -                & 0.454                  \\ 
ASTMNet \cite{li2022asynchronous}                          & 1D TCNN                             & CNN+RNN                            & SSD                                          & Yes                                & >100M                                  & 3                                    & -                & 0.467                  \\
MatrixLSTM \cite{Cannici2020Matrix}                          & LSTM                             & CNN                           & YOLOv3                                          & Yes                                & 61.5M                                  & -                                    & -                & 0.310                  \\

\hline
VC-DenseNet \cite{cordone2022object}                     & Voxel Cube                          & SNN                                & SSD                                          & Yes                                & 8.2M                                   & 5                                    & -                & 0.189                  \\
VC-MobileNet \cite{cordone2022object}                    & Voxel Cube                          & SNN                                & SSD                                          & Yes                                & 24.26M                                 & 5                                    & -                & 0.147                  \\
LT-SNN \cite{hasssan2023ltsnn} & HIST. & SNN & YOLOv2& Yes & - & - & -& 0.298 \\
KD-SNN \cite{bodden2024spiking} & HIST. & SNN & CenterNet& Yes & 12.97M & 5 & -& 0.229 \\
EMS-ResNet10  \cite{su2023deep}                   & HIST.                               & SNN                                & YOLOv3                                      & Yes                                & 6.2M                                   & 5                                    & 0.547            & 0.267                  \\
EMS-ResNet18  \cite{su2023deep}                   & HIST.                               & SNN                                & YOLOv3                                      & Yes                                & 9.34M                                  & 5                                    & 0.565            & 0.286                  \\
EMS-ResNet34 \cite{su2023deep}                    & HIST.                               & SNN                                & YOLOv3                                      & Yes                                & 14.40M                                 & 5                                    & 0.590            & 0.310                  \\
TR-YOLO  \cite{yuan2024trainable}                        & HIST.                               & SNN                                & YOLOv3                                      & Yes                                & 8.7M                                   & 3                                    & 0.451            & -                      \\ \hline
\multirow{3}{*}{\textbf{EAS-SNN (M)}}   & \multirow{3}{*}{\textbf{ARSNN}}     & \multirow{3}{*}{\textbf{SNN}}      & {YOLOX}$^\diamond$                 & {Yes}                       & 25.3M                                  & 3                                    & 0.731            & 0.409                  \\
                                 &                                     &                                    & {YOLOX}$^\ast$         & {Yes}                       & 25.3M                                  & 3                                    & 0.718            & 0.393                  \\
                                 &                                     &                                    & {YOLOX}$^\dagger$  & {Yes}                       & 25.3M                                  & 3                                    & 0.699            & 0.375                  \\ \hline
\multirow{3}{*}{\textbf{EAS-SNN (S)}}   & \multirow{3}{*}{\textbf{ARSNN}}     & \multirow{3}{*}{\textbf{SNN}}      & {YOLOX}$^\diamond$                & {Yes}                       & 8.92M                                  & 3                                    & 0.687            & 0.372                  \\
                                 &                                     &                                    & {YOLOX}$^\ast$            & {Yes}                       & 8.92M                                  & 3                                    & 0.692            & 0.372                  \\
                                 &                                     &                                    & {YOLOX}$^\dagger$ & {Yes}                       & 8.92M                                  & 3                                    & 0.675           & 0.354                  \\ \hline
                                 
                                 \multicolumn{9}{l}{ \it  $\dagger$ denotes the model uses a spiking backbone, a spiking feature pyramid network, and a spiking detection head.} \\
                                   \multicolumn{9}{l}{ \it  $\ast$ denotes the model uses a spiking backbone, a spiking feature pyramid network, and a non-spiking detection head.} \\
                                 \multicolumn{9}{l}{ \it  $\diamond$ denotes the model uses a spiking backbone, a non-spiking feature pyramid network, and a non-spiking detection head. } 

\end{tabular}
}

\end{table}
\begin{table}[t]
\centering
\caption{Performance comparison with the existing methods on N-Caltech 101.}
\label{tb.ncaltech}
\vspace{-8pt}
\resizebox{.95\linewidth}{!}{
    \begin{tabular}{lcccccccc}
    \toprule
                                      &                                      &                        &             &                                           &                                         &                                       & \multicolumn{2}{c}{}                                                \\
    \multirow{-2}{*}{\textbf{Method}} & \multirow{-2}{*}{\textbf{Representation}} & \multirow{-2}{*}{\textbf{Backbone}} & \multirow{-2}{*}{\textbf{Head}}& \multirow{-2}{*}{\textbf{Temporal}} & \multirow{-2}{*}{\textbf{\#Params}} & \multirow{-2}{*}{\textbf{Timestep}} & \multicolumn{2}{c}{\multirow{-2}{*}{\textbf{mAP$_{50}$ mAP$_{50:95}$}}} \\ \midrule
    YOLE \cite{cannici2019asynchronous}                              & Leaky Surface                        & CNN                                 & YOLO   & No                                     & -                                       & -                                     & 0.398                             & -                               \\
    AEGNN  \cite{schaefer2022aegnn}                           & Graph                                & GNN                                 & YOLO & No                                      & 20.0M                                    & -                                     & 0.595                             & -                               \\
    NVS-S \cite{li2021graph}                             & Graph                                & GNN                                 & YOLO & No                                      & 0.9M                                     & -                                     & 0.346                             & -                               \\
    ASYNet \cite{messikommer2020event}                         & HIST.                                & CNN                                 & YOLO  & No                                     & {\color[HTML]{333333} -}                & -                                     & 0.643                             & -                               \\
    ASYNet \cite{messikommer2020event}                          & QUE.                                 & CNN                                 & YOLO& No                                       & {\color[HTML]{333333} -}                & -                                     & 0.615                             & -                               \\ \midrule
    \textbf{EAS-SNN(M)}                            & \textbf{ARSNN}                                & \textbf{SNN}                                 & YOLO-X   & Yes                                 & 25.3M                                    & 3                                     & 0.664                             & 0.437                           \\
        \textbf{EAS-SNN(S)}                            & \textbf{ARSNN}                                & \textbf{SNN}                                 & YOLO-X     & Yes                                   & 8.92M                                    & 3                                     & 0.538                             & 0.338  \\               
    \bottomrule
    \end{tabular}
    % \vspace{-5pt}
}
\end{table}
\subsection{Performance Comparison}
\label{sc.pc}
We evaluate our proposed method against existing models on the Gen1, N-Caltech 101 and 1Mpx datasets.  To improve the overall efficacy,
we adopt PLIF \cite{fang2021incorporating} neurons with learnable decay and tdBN \cite{zheng2020going} in the downstream detector. The models, with sizes of 8.92M (SYOLOX-S) and 25.3M (SYOLOX-M) parameters, mirror the network architectures of YOLOX \cite{yolox2021}. As shown in Table \ref{tb.gen1}, our technique outperforms the state-of-the-art spike-based method \cite{su2023deep} (0.354 vs. 0.310 mAP$_{50:95}$) while using fewer parameters (8.92 M vs. 14.40 M) under three time steps. Even compared with ANNs that feature well-designed attention mechanisms \cite{gehrig2023recurrent} and recurrent architectures \cite{li2022asynchronous,perot2020learning}, our method remains competitive and notably achieves over 40\% mAP$_{50:95}$ on Gen1 through sparse spike-driven operations, thereby enhancing energy efficiency as detailed in \cref{sc.ee}. Regarding N-Caltech 101 (Table \ref{tb.ncaltech}), our approach improves the detection performance with spike-based representation. On the 1Mpx dataset, the largest event-based detection dataset, the proposed method achieves the first spike-based results, demonstrating remarkable performance using SNNs, as depicted in Table \ref{tb.gen4}.

% In object detection, modules are designed for specific roles: backbones extract features, Feature Pyramid Networks (FPN) merge multi-scale features, and detection heads handle coordinate regression and object recognition. We examined sparse spike representation effects by integrating a non-spiking detection head and FPN into a hybrid architecture. As shown in Table \ref{tb.gen1}, a spiking FPN within a compact 8.92M network outperforms its non-spiking counterpart, indicating spike-based architectures might offer regularization advantages for event-driven detection. However, spiking detection heads within constrained time steps appear to reduce network performance.

In object detection, each module has a specific role: backbones extract features, Feature Pyramid Networks (FPN) fuse multi-scale features, and detection heads handle coordinate regression and object recognition. We explored the impact of sparse spike representation by integrating a non-spiking detection head and FPN into a hybrid architecture. As shown in Table \ref{tb.gen1}, a spiking FPN within a compact 8.92M network outperforms its non-spiking counterpart, indicating that spike-based architectures might offer regularization benefits for event-based detection. However, spiking detection heads within constrained time steps appear to reduce network performance.

\begin{table}[t]
\caption{Performance comparison with the state-of-the-art methods on 1Mpx. 
% The letters M and S in parentheses denote medium and small-scale network structures, respectively.
 }
\label{tb.gen4}
\vspace{-8pt}
\resizebox{\linewidth}{!}{
\begin{tabular}{@{}lcccccccc@{}}
\hline
\multirow{2}{*}{\textbf{Method}} & \multirow{2}{*}{\textbf{Representation}} & \multirow{2}{*}{\textbf{Backbone}} & \multirow{2}{*}{\textbf{Head}}               & \multirow{2}{*}{\textbf{Temporal}} & \multirow{2}{*}{\textbf{\#Params}} & \multirow{2}{*}{\textbf{Timestep}} &   \multirow{2}{*}{\textbf{$\text{mAP}_{50}$}} &\multirow{2}{*}{\textbf{$\text{mAP}_{50:95}$}}       \\ 
                                 &                                     &                                    &                                              &                                    &                                       &                                      &  &  \\ \hline
RED \cite{perot2020learning}                              & Event Volum                         & CNN+RNN                            & SSD                                          & Yes                                & 24.1M                                 & -                                    & -                & 0.430                    \\
RVT \cite{gehrig2023recurrent}                             & HIST.                               & Transformer+RNN                  & YOLOX                                        & Yes                                & 18.5M                                & 21                                   & -                & 0.474            \\
ERGO-12 \cite{zubic2023chaos} & ERGO-12 & Transformer & YOLOv6& No & - & - & -& 0.406 \\
ERGO-12 \cite{zubic2023chaos} & HIST. & Transformer & YOLOv6& No & - & - & -& 0.327 \\

E2VID
 \cite{rebecq2019high}                          & Reconstruction                            & CNN                            & RetinaNet                                          & No                                & -                                  & -                                    & -                & 0.250                  \\
 ASTMNet \cite{li2022asynchronous}                          & 1D TCNN                             & CNN+RNN                            & SSD                                          & Yes                                & >100M                                  & 3                                    & -                & 0.483                  \\

Events+YOLOv3 \cite{jiang2019mixed}                          & HIST.                             & CNN                            & YOLOv3                                          & No                                & -                                  & -                                    & -                & 0.346                  \\
% ASTMNet \cite{li2022asynchronous}                          & 1D TCNN                             & CNN+RNN                            & SSD                                          & Yes                                & >100M                                  & 3                                    & -                & 0.483                  \\
% ASTMNet \cite{li2022asynchronous}                          & 1D TCNN                             & CNN+RNN                            & SSD                                          & Yes                                & >100M                                  & 3                                    & -                & 0.483                  \\
\midrule
            
{\textbf{EAS-SNN (M)}}   & {\textbf{ARSNN}}     & {\textbf{SNN}}      & {YOLOX}                & {Yes}                       & 25.3M                                  & 3                                    & 0.651            & 0.362                  
                                                               \\ \hline

\end{tabular}
}

\end{table}

\subsection{Experimental Analysis}
\label{sc.ea}
\noindent {\footnotesize \textbf{{Scalability to Dense Neural Network:}} } 
\begin{figure}[t]
    \centering
    \includegraphics[width=.87\linewidth]{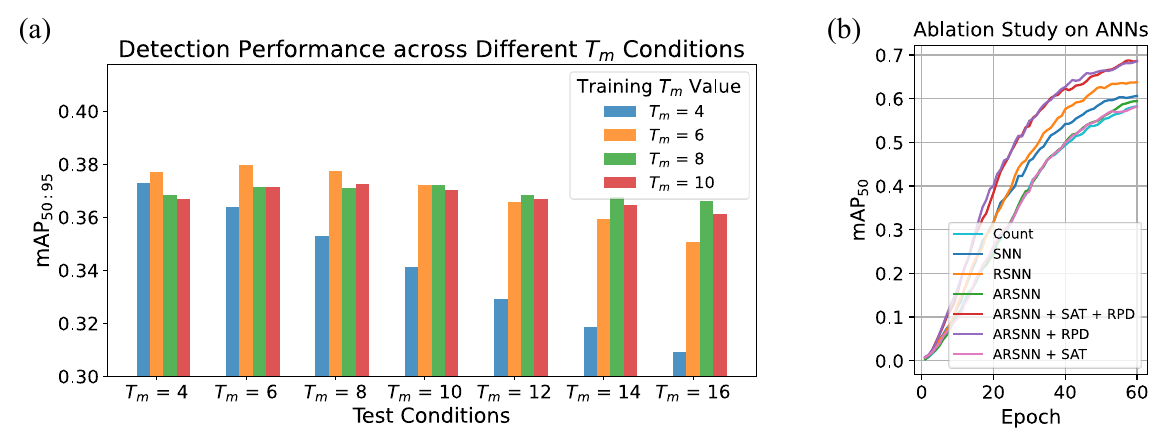}
    \vspace{-5pt}
    \caption{(a) Assessing the scalability of the early aggregation step $T_m$ in mismatched testing scenarios on Gen1. (b) Evaluating the effectiveness of the proposed adaptive sampling mechanism within conventional dense neural networks on N-Caltech 101.    }
        % \vspace{-5pt}
    \label{fig.ann_ab}
\end{figure}
In this section, our objective is to validate the efficacy of the proposed adaptive sampling technique within conventional dense neural networks. Here, we only transmit the initial aggregated potential to subsequent networks. \cref{fig.ann_ab}b displays the mAP$_{50}$ training curves on the N-Caltech 101 dataset, comparing different methods. The combination of ARSNN with SAT and RPD demonstrates promising performance improvements, while ARSNN with SAT or RPD alone leads to declined performance, highlighting the synergistic benefits of our proposed modules.  Additionally, recurrent synaptic connections further improve network performance.
% These results validate the adaptability of our sampling module for dense neural networks.
% showcasing its potential to significantly improve neural network performance.
% In this part, we aim to verify the proposed adaptive sampling method on typical dense neural networks.  To achieve this, each neuron in the last layer of the embedding module only sends its potential aggregation of the first sample into the downstream networks. \cref{fig.ann_ab} shows the training curve of mAP$_{50:95}$ with different methods on the N-Caltech 101 dataset. The method of ARSNN+SAT+RPD achieves significant performance improvement while the ARSNN+SAT and ARSNN+RPD bring severe performance degradation, which further validates the synergistic effect of the proposed modules. With recurrent synaptic connection, the performance of spike-based embedding has also been improved. These results show the whole framework of adaptive sampling can be integrated into dense neural networks. 

% 对每一个采样得到的
\begin{figure}[t]
    \centering
    \includegraphics[width=.88\linewidth]{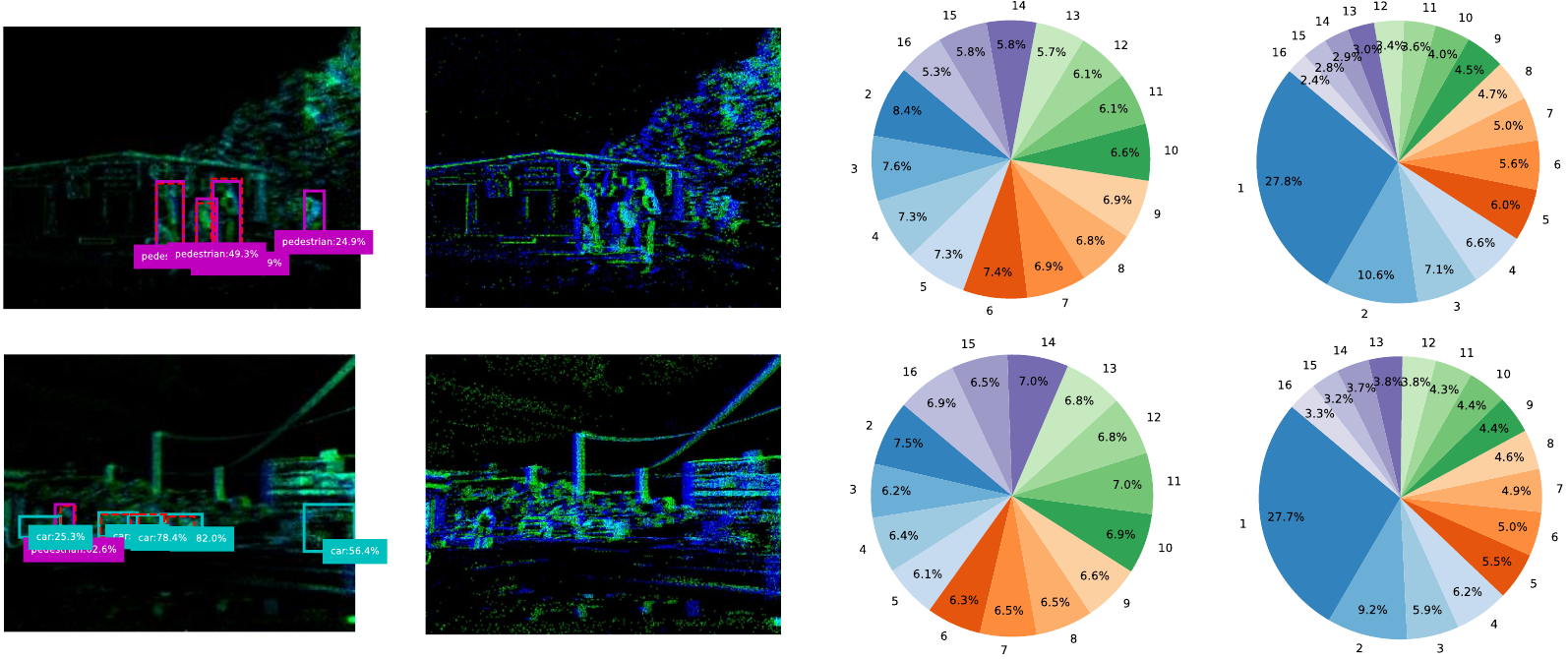}
    \vspace{-10pt}
    \caption{Exploring the behavior of sampling module. The first column illustrates bounding boxes of prediction, whereas the second column provides pixel-wise firing count of spike neurons.  The third and fourth columns, respectively, depict the distribution of neuron firing time and the sampling duration, quantified as the inter-spike interval.  }
    \label{fig.slice}
    % \vspace{-5pt}
\end{figure}
\begin{figure}[t ]
    \centering
    \includegraphics[width=.88\linewidth]{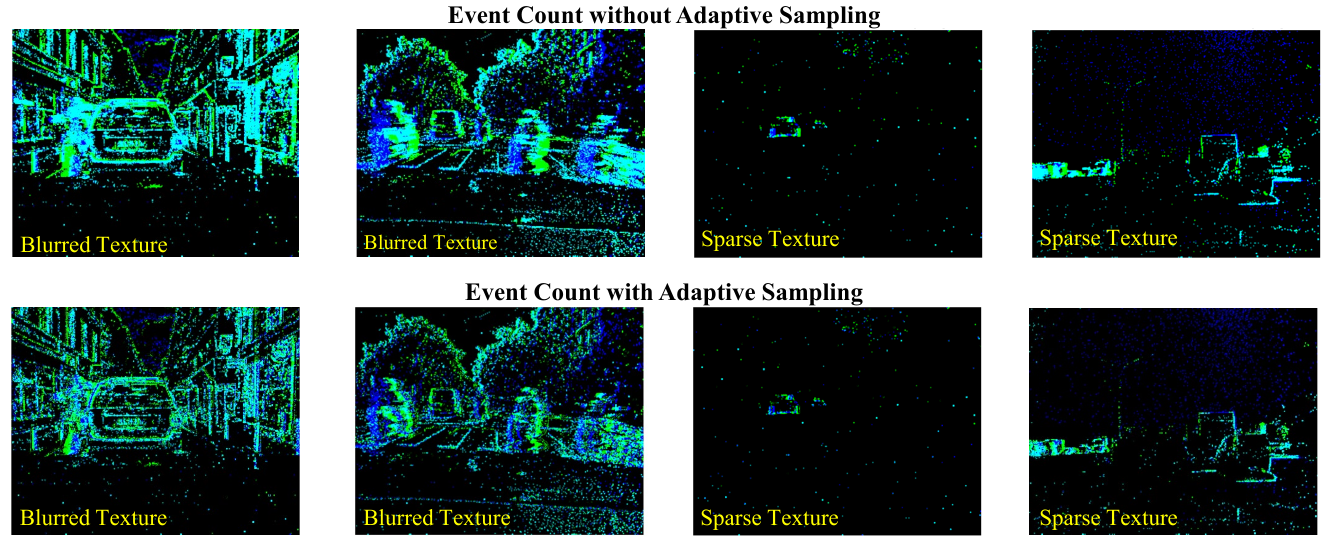}
        \vspace{-5pt}
    \caption{ Comparison of event count images with and without the adaptive sampling module under blurred and sparse contexts. }
    \label{fig.count_samp}
            % \vspace{-15pt}
\end{figure}
\noindent {\footnotesize \textbf{{Impact of Early Aggregation:}} } We investigate the influence of early aggregation, as defined in \cref{eq.count}, by modulating the aggregation step $T_m$ on Gen1 using spiking YOLOX-S. The first three slices sampled from $T_m$-step inputs are aggregated and fed sequentially into a spike-based backbone. The parameter $T_m$ indicates the temporal resolution of inputs, where higher $T_m$ corresponds to finer granularity. As depicted in \cref{fig.ann_ab}a, optimal performance is generally observed when $T_m$ remains consistent between the training and testing phases. Notably, the model trained with $T_m=4$, sustains over 30\% mAP$_{50:95}$, despite a fourfold increase in granularity. Overall, models trained with $T_m=8$ demonstrate considerable robustness, delivering promising performance under all tested conditions.

\noindent {\footnotesize \textbf{{Distribution of Event Sampling:}} } In this part, we employ the model trained under $T_m=8$ and examine the behavior of each neuron while using $T_m=16$ for inference. The sampling intervals of various spiking neurons are shown in the last column of \cref{fig.slice}, which demonstrates significant variability. The longest sample covers the full input, while the shortest sample only includes the input of one time step. Here, we disregard the neurons without spike firing, which typically make up $80\%+$ due to the sparse event input. Additionally, compared to the event count representation in the first column, the image of the firing count in the second column offers a clearer estimation of the scene. The qualitative results of adaptive sampling are further presented in  \cref{fig.count_samp}. The top row shows event count image without adaptive sampling, under blurred and sparse textures. The bottom row demonstrates the improvement in information retention and noise reduction with adaptive sampling in the same contexts.

\subsection{Energy Efficiency and Inference Speed}
\label{sc.ee}

While ANNs execute dense multiplication-and-accumulation (MAC) operations determined by their architecture, SNNs on neuromorphic hardware utilize input-conditioned spikes for sparse accumulation (AC) operations, thereby enhancing energy efficiency. 
To evaluate the energy efficiency of SNNs, we compute the average AC operations across Gen1, as detailed in Table \ref{tb.energy}. According to \cite{han2015learning}, we estimate the energy cost for 32-bit floating-point AC and MAC as $0.9 \ pJ$ and $4.6 \ pJ$ per operation, respectively. 
The results show a significant reduction in energy consumption with spike-driven computation, achieving a $5.85\times$ reduction for YOLOX-M and $3.73\times$ for YOLOX-S.  It is worth noting that, for ANNs, we omit the energy consumption of the embedding phase, offering a comparatively conservative estimate of power savings. Detailed firing rates and additional results for different models are provided in Supplementary Materials.

Furthermore, we compared the frames per second (FPS) performance of our model with other state-of-the-art methods on the Gen1 dataset. Our model achieves competitive detection speeds of 54.35 FPS, compared to 59.88 FPS for RED \cite{perot2020learning}, 28.09 FPS for ASTMNet \cite{li2022asynchronous}, 106.38 FPS for RVT-T \cite{gehrig2023recurrent}, and 59.52 FPS for GET \cite{peng2023get}. This demonstrates the significant potential for higher efficiency on neuromorphic chips, which support sparse spike-driven operations.

\begin{table}[t]
\centering
\caption{Comparison of energy consumption based on number of synaptic operations.}
\vspace{-6pt}
\label{tb.energy}
\resizebox{\linewidth}{!}{
\begin{tabular}{|c|cc|ccc|cc|ccc|}
\hline
\multirow{2}{*}{\textbf{Module}} & \multicolumn{2}{c|}{\textbf{YOLOX-M}}           & \multicolumn{3}{c|}{\textbf{Spiking YOLOX-M}}                                  & \multicolumn{2}{c|}{\textbf{YOLOX-S}}           & \multicolumn{3}{c|}{\textbf{Spiking YOLOX-S}}                                  \\ \cline{2-11} 
                                 & \multicolumn{1}{c|}{MAC(G)} & Energy(mJ)        & \multicolumn{1}{c|}{AC(G)} & \multicolumn{1}{c|}{MAC(G)} & Energy(mJ)          & \multicolumn{1}{c|}{MAC(G)} & Energy(mJ)        & \multicolumn{1}{c|}{AC(G)} & \multicolumn{1}{c|}{MAC(G)} & Energy(mJ)          \\ \hline
\textbf{Embedding}               & \multicolumn{1}{c|}{0}      & 0                 & \multicolumn{1}{c|}{0.02}  & \multicolumn{1}{c|}{1.63}   & 7.52                & \multicolumn{1}{c|}{0}      & 0                 & \multicolumn{1}{c|}{0.02}  & \multicolumn{1}{c|}{1.63}   & 7.52                \\
\textbf{Backbone}                & \multicolumn{1}{c|}{16.21}  & 74.57             & \multicolumn{1}{c|}{8.49}  & \multicolumn{1}{c|}{1.41}   & 14.12               & \multicolumn{1}{c|}{5.24}   & 24.10             & \multicolumn{1}{c|}{2.66}  & \multicolumn{1}{c|}{0.71}   & 5.64                \\
\textbf{FPN}                     & \multicolumn{1}{c|}{8.19}   & 37.68             & \multicolumn{1}{c|}{3.40}  & \multicolumn{1}{c|}{0}      & 3.06                & \multicolumn{1}{c|}{2.63}   & 12.09             & \multicolumn{1}{c|}{1.26}  & \multicolumn{1}{c|}{0}      & 1.14                \\
\textbf{Head}                    & \multicolumn{1}{c|}{11.33}  & 52.10             & \multicolumn{1}{c|}{3.78}  & \multicolumn{1}{c|}{0}      & 3.40                & \multicolumn{1}{c|}{5.04}   & 23.17             & \multicolumn{1}{c|}{1.79}  & \multicolumn{1}{c|}{0}      & 1.61                \\ \hline
\textbf{Total}                   & \multicolumn{1}{c|}{35.73}  & 164.35(1$\times$) & \multicolumn{1}{c|}{15.68} & \multicolumn{1}{c|}{3.04}   & 28.10(5.85$\times$) & \multicolumn{1}{c|}{12.90}  & 59.35(1$\times$) & \multicolumn{1}{c|}{5.74}  & \multicolumn{1}{c|}{2.34}   & 15.91($3.73\times$) \\ \hline
\end{tabular}
% \vspace{-25pt}

% \vspace{-10pt}

% \vspace{-28pt}
}

\end{table}
\section{Conclusion}
We introduce an adaptive sampling method utilizing recurrent spiking neural networks (ARSNN), enhanced by Residual Potential Dropout (RPD) and Spike-Aware Training (SAT) specifically to mitigate performance degradation in event-based sampling. Integrated with a spike-based architecture in the downstream neural network, our comprehensive framework, EAS-SNN, outperforms existing spike-based detection methods and offers superior energy efficiency compared to ANNs. To the best of our knowledge, it is the first effort toward end-to-end optimization for event-based sampling and representation. Additionally, we demonstrate the versatility of our proposed sampling technique through its application to dense neural networks.  Despite exploring robustness across varying granularities of early aggregation, the ultimate goal 
is to develop a spike-based sampler that can precisely synchronize with the temporal granularity of raw event streams, measured in microseconds, without necessitating early aggregation. Our future work will focus on achieving this granularity, pending advancements in neuromorphic computing infrastructure.
% ncements in neuromorphic computing infrastructure.
% ********************AAAAABBBBBB***********CCCDEFG 

% \subsection{Acknowledgement}
 \section*{Acknowledgements} This work was supported by National Natural Science Foundation of China under Grant 62236007, and the Key R\&D Program of Zhejiang under Grant 2023C03001 and 2022C01048.

\bibliographystyle{splncs04}
\bibliography{main}
\end{document}